\documentclass[letterpaper, 10 pt, journal, twoside]{ieeetran}

\IEEEoverridecommandlockouts                              

\usepackage{graphics} 
\usepackage{epsfig} 
\usepackage{mathptmx} 
\usepackage{times} 
\usepackage{amsmath} 
\usepackage{amssymb}  
\usepackage{algorithm}
\usepackage{graphicx}
\usepackage{algpseudocode}
\usepackage{tabularx}
\usepackage{soul}
\usepackage{environ}
\usepackage{multirow}
\usepackage{url}
\usepackage{color, colortbl}
\definecolor{Gray}{gray}{0.9}
\usepackage[table]{xcolor} 
\usepackage{threeparttable}
\usepackage{booktabs}
\usepackage[T1]{fontenc}
\usepackage[utf8]{inputenc}
\usepackage{babel}
\usepackage{cite}

\newcommand{\fref}[1]{Fig.~\ref{#1}}

\begin{document}

\title{Robotic Depowdering for Additive Manufacturing Via Pose Tracking}

\author{Zhenwei Liu,
Junyi Geng,
Xikai Dai,
Tomasz Swierzewski,
and Kenji Shimada
\thanks{Zhenwei Liu, Xikai Dai, Tomasz Swierzewski, and Kenji Shimada are with the Department of Mechanical Engineering, Carnegie Mellon University, Pittsburgh, PA 15213 USA (emails: zhenweil.career@gmail.com; \{xikaid, tswierze, shimada\}@andrew.cmu.edu).}
\thanks{Junyi Geng is with the Robotics Institute, Carnegie Mellon University, Pittsburgh, PA 15213 USA (email: junyigen@andrew.cmu.edu).}
}

\markboth{IEEE Robotics and Automation Letters. Preprint Version. Accepted July, 2022}
{Liu \MakeLowercase{\textit{et al.}}: Robotic Depowdering} 

\maketitle

\begin{abstract}
With the rapid development of powder-based additive manufacturing, depowdering, a process of removing unfused powder that covers 3D-printed parts, has become a major bottleneck to further improve its productiveness. Traditional manual depowdering is extremely time-consuming and costly, and some prior automated systems either require pre-depowdering or lack adaptability to different 3D-printed parts.
To solve these problems, we introduce a robotic system that automatically removes unfused powder from the surface of 3D-printed parts.
The key component is a visual perception system, which consists of a pose-tracking module that tracks the 6D pose of powder-occluded parts in real-time, and a progress estimation module that estimates the depowdering completion percentage. The tracking module can be run efficiently on a laptop CPU at up to 60 FPS. 
Experiments show that our system can remove unfused powder from the surface of various 3D-printed parts without causing any damage. To the best of our knowledge, this is one of the first vision-based depowdering systems that adapt to parts with various shapes without the need for pre-depowdering.

\end{abstract}


\section{Introduction}

\begin{figure}
\begin{center}
\setlength{\abovecaptionskip}{0pt}
\includegraphics[width=1.0\linewidth]{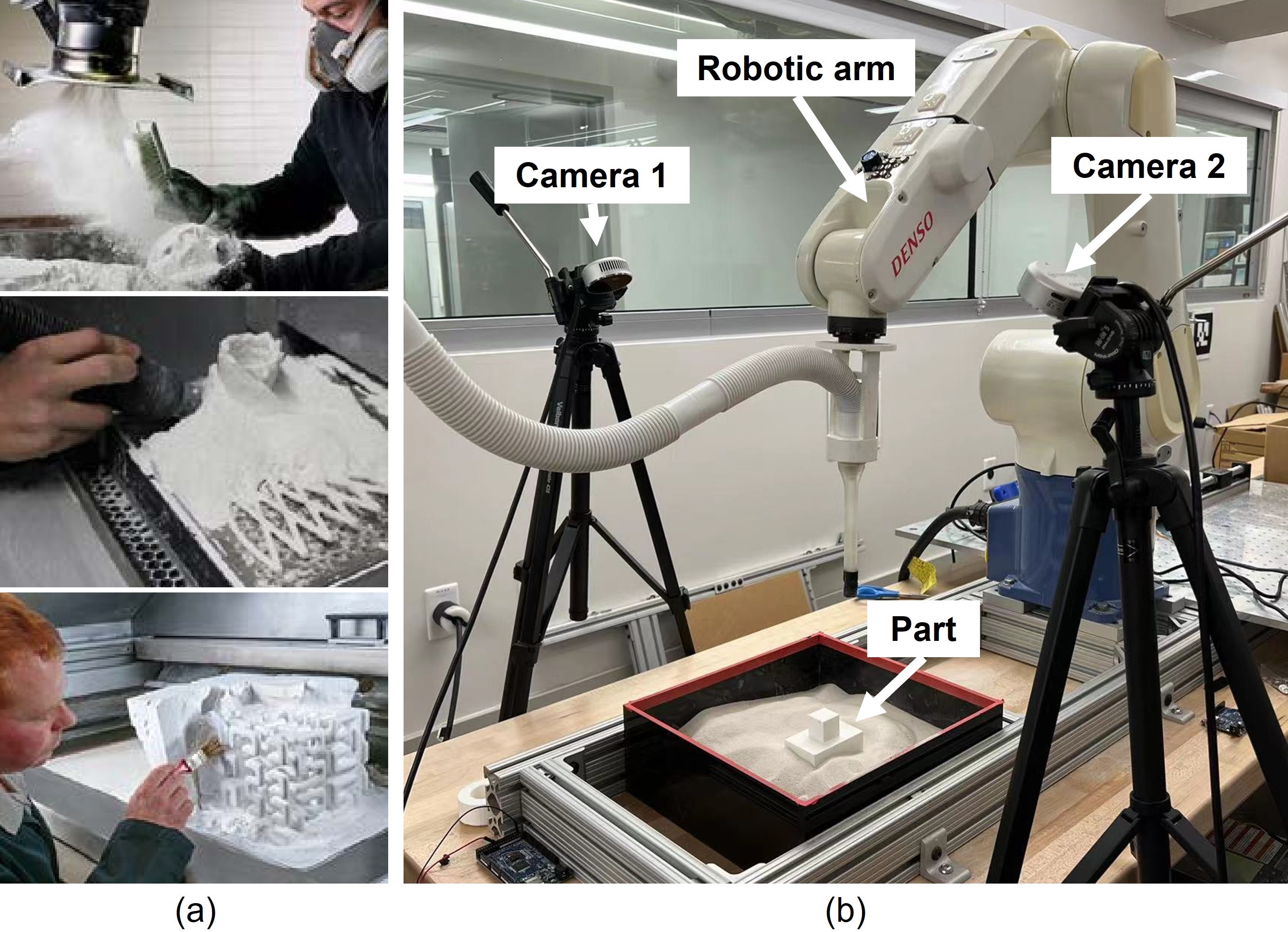}
\end{center}
\vspace{-10pt}
   \caption{(a) Examples of manual depowdering. Human operators use blowers, vacuums, and brushes to remove the unfused powder. (b) Overview of our depowdering system. The point cloud sequence from 3D cameras is fed into the visual perception system to track the 6D pose of the powder-occluded part. Based on the estimated pose, a depowdering path is generated for the robot to remove powder through vacuuming and air blasting.}
   \vspace{-10pt}
\label{fig:manual_depowdering}
\end{figure}

\IEEEPARstart{D}{epowdering}, a process of removing unfused powder surrounding 3D-printed parts, is an important post-processing step for powder-based additive manufacturing. For example, 3D-printed parts need to be extracted from the build box with the residual powder removed, before being sent to subsequent processes such as heat treatment and surface finishing. 
Traditionally, depowdering is often performed manually, where the human operators use shovels, vacuums, brushes, and blowers to remove powder incrementally, as shown in \fref{fig:manual_depowdering} (a). However, this process is tedious and time-consuming, resulting in high operational costs. In addition, the airborne powder raised during depowdering may cause damage to the human respiratory system. Therefore, the automation of depowdering has become an urgent need.

Some researchers developed simple automated systems to tackle the depowering problem. 
For example, the HP automatic unpacking station \cite{HpJetFusion} and Solukon SFP series \cite{SolukonPlastic} are able to perform depowdering for small-to-medium sized parts through a combination of vibration, vacuum, and air blasting. However, these solutions may damage fragile parts, such as delicate polymer parts and unsintered metal parts in binder-jet printing. The process also results in unknown part poses, which raises the difficulty for subsequent part handling. While some researchers directly use a compliant gripper to extract parts from the unfused powder \cite{AutomatedExtraction}, the method may not work well for large parts, since the drag force generated when an object moves through powder increases with the object size\cite{Albert2001GranularDO}. Besides, some fragile parts may be unsuitable for gripping.

Other automated systems aim to remove the small amount of residual powder that remains on/inside 3D-printed parts. The Solukon SFM series \cite{SolukonMetal} rotate the entire part in different orientations to let the powder flow out of the inner structure. However, it requires 3D-printed parts to be pre-depowdered. Nguyen et el. \cite{Decaking} used MaskRCNN \cite{maskrcnn} to localize 3D-printed parts and rub them on a brush to remove powder that sticks to the surface. 
Although the system uses vision feedback for part localization, the 3D-printed parts still need to be pre-depowdered.
In addition, it only handles flat shapes and requires a dataset of each specific part for neural network training, which is difficult to obtain for customized parts.

\begin{figure}[h]
\begin{center}
\includegraphics[width=1.0\linewidth]{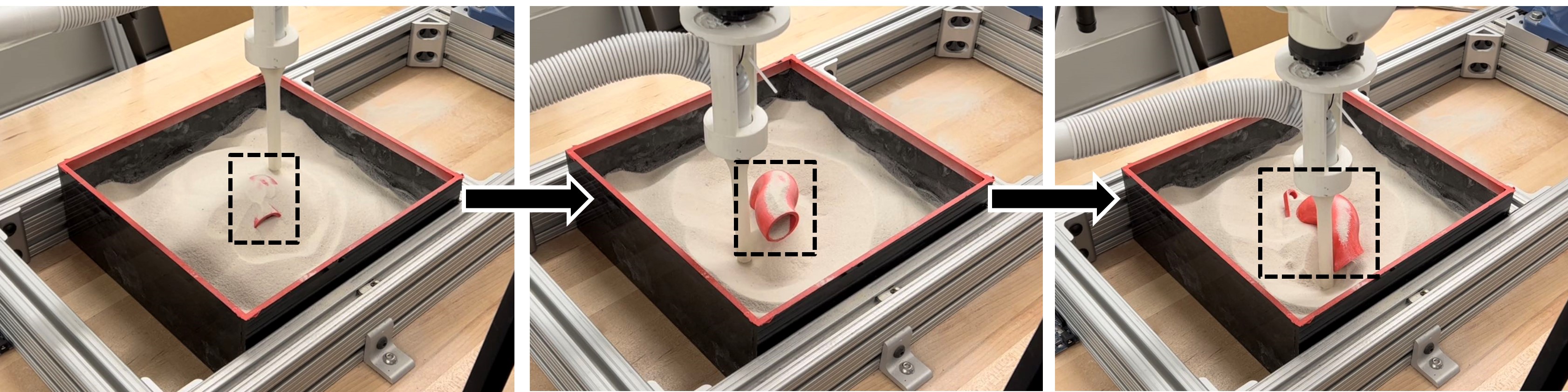}
\end{center}
\vspace{-10pt}
\caption{Example of a 3D-printed part losing balance during the depowdering process. As the vacuum constantly removes powder, the part gradually loses powder support and finally falls to one side.}
\vspace{0pt}
\label{fig:lose_balance}
\end{figure}

To solve the above problems, we resort to vision-based 6D pose tracking.
However, directly applying existing pose-tracking methods is challenging due to various reasons. For instance, 3D-printed parts appear in a similar color to the surroundings since they consist of the same material. Due to various part shapes and powder occlusion, features such as edges may not be available. In addition, the visible surface of 3D-printed parts gradually changes during depowdering, which poses extra challenges. In scenarios shown in \fref{fig:lose_balance}, where 3D-printed parts are not anchored to the build box, they might move or lose balance when touched by cleaning tools or when  powder support is removed. These challenges indicate that the vision system needs to be carefully designed.

\vspace{-5pt}

In this paper, we develop a vision-based robotic depowdering system that automatically removes unfused powder from the surface of various types of 3D-printed parts without the need for pre-powdering,
as shown in \fref{fig:manual_depowdering} (b). 
The system consists of two major components: a visual perception system (VPS) and a motion planning system (MPS). 
This paper mainly focuses on the development of VPS, which is to track the 6D poses of powder occluded parts in real-time, identify the powder contour around parts, and estimate the depowdering progress.
To demonstrate the effectiveness of our VPS, we apply a simple motion planner to generate a depowdering path for the robotic arm based on the part pose and the powder contour provided by VPS. 

In particular, we design an algorithm that leverages iterative closest point (ICP) \cite{ICP} and template update \cite{TemplateUpdate} to track the part pose. It constantly registers a template to the point cloud scan, with the template derived from the CAD model based on the current visual appearance of the part. In addition, we propose a conditional template update strategy to avoid accumulated template update error. We then extract the powder contour around the 3D-printed part and estimate the depowdering progress based on the 6D pose. Our approach does not rely on feature extraction, thus is robust to point cloud noise and occlusion. Different from other learning-based methods that require large dataset and neural network training, our method, due to the nature of template matching, only requires the CAD model and the initial object pose of the 3D-printed part, which are usually available in additive manufacturing.

To summarize, the main contributions of this paper are:
\begin{enumerate}
\item We develop a vision-based robotic system that automatically removes unfused powder from the surface of 3D-printed parts, which is applicable to different shapes and avoids damaging fragile parts.
To the best of our knowledge, this is one of the first vision-based depowdering systems that adapt to different part shapes without the need for pre-depowdering.

\item We design an efficient pose-tracking algorithm, which combines ICP and conditional template update without the need for a large dataset or neural network training.

\item We present thorough experiments to evaluate the tracking performance on various 3D-printed parts. The tracking pipeline is computationally efficient, which can be run on a laptop CPU at maximum 60 FPS.
\end{enumerate}
\vspace{-5pt}

\section{Related Work}
\subsection{Automated Depowdering}
Traditionally, depowdering tasks are accomplished by human labor. However, the huge demand due to the recent growth of 3D-printing technologies raises the need for automated depowdering. As mentioned previously, prior works address automated depowdering through a combination of vibration, vacuum, and air blasting\cite{HpJetFusion}\cite{SolukonPlastic}, using a compliant gripper to directly extract 3D-printed parts \cite{AutomatedExtraction}, rotating parts in different orientations \cite{SolukonMetal}, and rubbing parts on a brush based on visual localization \cite{Decaking}. Compared with previous works, our robotic system is applicable to 3D-printed parts with different shapes, does not require 3D-printed parts to be pre-depowdered, and, due to the nature of template matching, does not need a large dataset for neural network training. 

\begin{figure*}
\begin{center}
\includegraphics[width=1.0\linewidth]{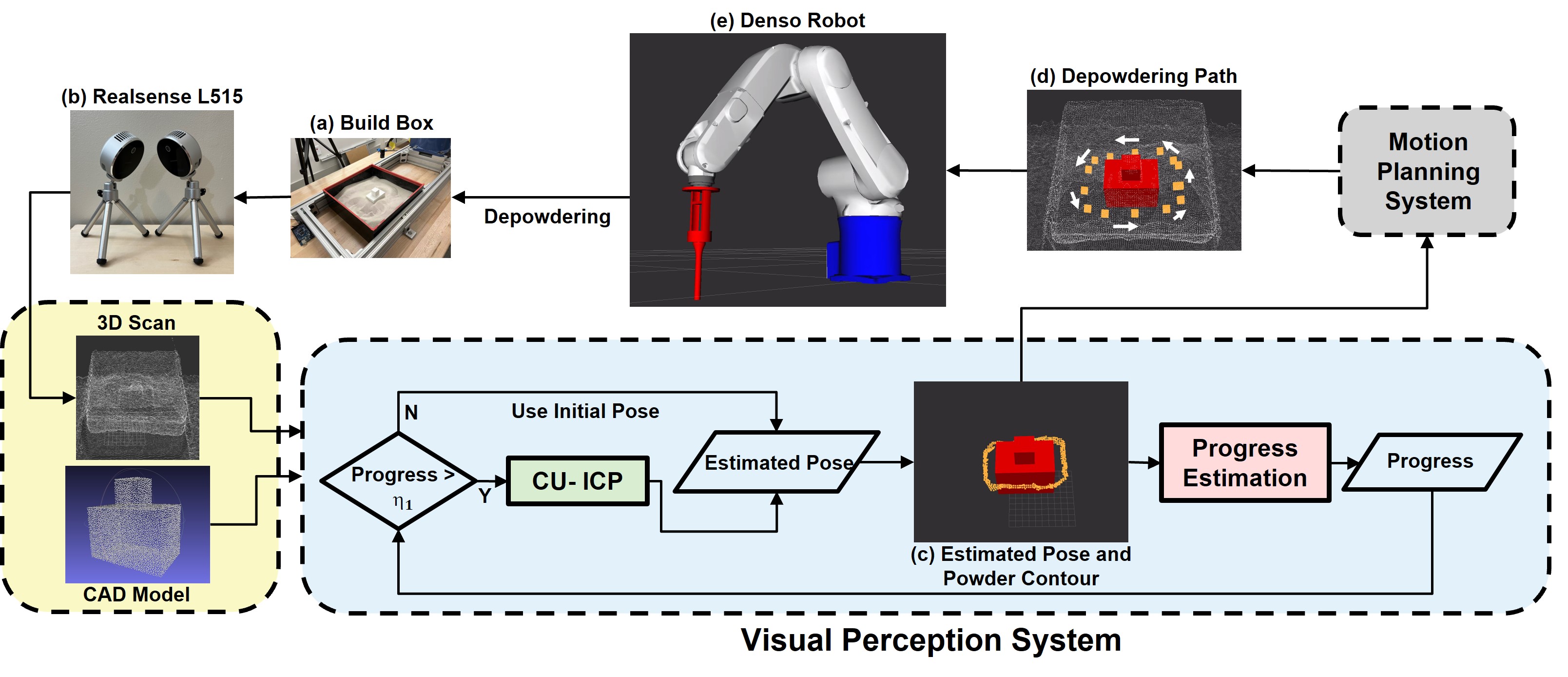}
\end{center}
\vspace{-20pt}
\caption{The architecture of our system. The inputs to the system are the CAD model and the point cloud can. When the depowdering progress is less than $\eta_1$, no tracking is performed, and the initial pose is used. Once the progress exceeds $\eta_1$, CU-ICP starts running. The robot first removes powder through vacuuming, and then finishes up depowdering by air blasting.}
\vspace{0pt}
\label{fig:architecture}
\end{figure*}

\subsection{6D Pose Estimation}
6D pose estimation provides critical information for robots to perform planning and execution. Current methods regarding 6D pose estimation can be categorized as \textit{instance-level}\cite{labbe2020, PoseRBPF,PoseCNN, wang2019densefusion, li2017deepim,SSD6D}, where the CAD models of known instances are given as priors, or \textit{category-level}\cite{ICaps,BundleTrack,6PACK,FSNet,CanonicalCategory,WangNormalizedCategory}, where the CAD model is not available and the target object may be unseen during training. Depending on whether deep learning is applied, 6D pose estimation can also be classified as \textit{geometry-based} \cite{CollectPoseRegistration,ChoiPoseDailyObject,ChoiVotingBased,ChoiTextureless,DrummondRealTimeTracking,PauwelsPoseEstimation} or \textit{learning-based}\cite{labbe2020, PoseCNN, 6PACK,FSNet,CanonicalCategory}. Each category has its own pros and cons. For example, learning-based methods can generalize to different backgrounds and lighting conditions.
However, these methods require a large dataset for neural network training, which may not be efficient enough for some applications such as depowdering.

Geometry-based methods, on the other hand, can be quickly implemented and deployed, and do not require a large dataset for training, which is highly suitable for depowdering. However, it usually utilizes well-selected features, which may not be robust for some featureless parts in depowdering. Iterative closest point (ICP) \cite{ICP}, a well-known geometry-based algorithm for point cloud registration, estimates the 6D pose by aligning the source point cloud to the target point cloud. While several variants of ICP have been proposed to improve the registration performance\cite{Fitzgibbon2001RobustRO,GoICP, SharpICPInvariant}, they either require feature selection or cannot achieve real-time performance. In this paper, we develop an adaptive algorithm that leverages ICP and template update. The algorithm does not rely on feature extraction, and is robust to sensor noise and occlusion.
\vspace{0pt}

\section{System Overview}
\subsection{Hardware Platform}
The main components of our hardware platform are:
\begin{itemize}
    \item a Denso VS-6577 6-axis manipulator, with a cleaning nozzle attached to the end-effector.
    \item two Realsense L515 cameras, which generate a point cloud sequence of the scanned object.
    \item a build box with 3D-printed parts and powder\footnote{We use children-play sand in the experiment since it is safe and easy to handle while mimics the property of the actual polymer and metal powder.} inside.   
\end{itemize}

\subsection{System Architecture}
\label{sec:architecture}
The architecture of the depowdering system is shown in Fig.~\ref{fig:architecture}. It consists of a visual perception system (VPS), which is the main focus of this paper, and a motion planning system (MPS). VPS tracks the part pose in real-time, extracts the powder contour around parts, and estimates the depowdering progress. The depowdering progress, $\eta$, is defined as the height ratio between the visible portion of the part and the complete part, which will be further explained in Sec.~\ref{sec:progress}. 
Then, based on the estimated pose and the powder contour, as shown in Fig.~\ref{fig:architecture} (c), MPS generates a depowdering path\footnote{The path is collision-free because it is always certain distances away from the outer contour of the visible part surface.} along the outer contour of the visible part surface, shown in Fig.~\ref{fig:architecture} (d). Overall, the depowdering process can be divided into three phases based on the progress $\eta$: 
\newline\indent
\textit{Phase 1:} $0 < \eta \leq \eta_1$. No pose tracking is performed because not enough part surface is available for tracking. 3D-printed parts are assumed to stay at the initial pose\footnote{During this stage, the majority of the 3D-printed part is still buried under powder. Therefore, the powder will hold the part in place.}. The powder contour and the path are generated based on the initial pose. The robotic arm follows the path and removes powder through vacuuming. We empirically select $\eta_1=30\%$ based on the observation from the depowdering experiments.
\newline\indent
\textit{Phase 2:} $ \eta_1 < \eta \leq \eta_2$. Pose tracking starts. The powder contour and depowdering path are generated based on the estimated part pose. The robot follows the path and removes powder through vacuuming. Note that the generated path automatically adjusts to a new pose when the part moves. We select $\eta_2=85\%$ as we observe that parts usually appear to be fully uncovered with progress larger than 85\%.
\newline\indent
\textit{Phase 3:} $\eta_2 < \eta \leq 100\%$. The majority of the 3D-printed part has been uncovered. To finish up depowdering, the robot goes over the entire part surface and removes the residual powder through air blasting from the top\footnote{Currently the vacuum is manually switched to air blower by reconnecting the hose. The power source switching can be automated in the future work.}. After this step, parts can be sent for further post-processing.

\section{Visual Perception System}
The 3D-printed parts are completely covered by unfused powder initially, with the powder surface slightly higher than the top of the part.
The initial part pose $T_{init}$ can be obtained from the 3D-printing system. With the CAD model $P_{cad}$, the initial pose $T_{init}$, and a sequence of point cloud scan $P_{i}, i = 1,...,t$ as input, the goals of the VPS are to:
\begin{itemize}
\item track the current 6D part pose $T_{i}$ in real-time. 
\item identify the powder contour around the part and estimate the depowdering progress based on $T_{i}$.
\end{itemize}

\vspace{0pt}
\subsection{Mismatch Dilemma in ICP}
\label{sec:mismatch}
With the known initial pose and the CAD model, we formulate the pose-tracking problem as a real-time registration problem. Given an initial pose $T_{init}$, the traditional ICP is able to align a source point cloud to a target point cloud by minimizing the $L_2$ error without relying on feature extraction. For our pose-tracking task, the source is the template point cloud $P_{tmpl}$ derived from the CAD model, and the target is the current point cloud scan $P_{i}$. By constantly aligning $P_{tmpl}$ to $P_{i}$ with the previous pose estimate $T_{i-1}$ as a prior, ICP outputs a relative transformation $T_{i,i-1}$ between the current pose and the previous pose. The new pose $T_{i}$ can then be derived by $T_{i} = T_{i,i-1} \times T_{i-1}$.
\newline\indent
One major challenge for ICP is the target appearance variation, i.e., parts start from fully occluded to fully visible. A common practice to address this is to update the template based on the current object appearance. While many researchers investigated the template update strategies \cite{TemplateUpdate}, most of them only address the scenario for 2D images. We explore further by extending these strategies from 2D to 3D. 
\newline\indent\textbf{Strategy 1} A naive way is to use the entire CAD model as a template and does not update the template. However, this could result in a mismatch between the template and the target due to the partially visible appearance, see Fig. 4 (c).
\newline\indent\textbf{Strategy 2} One way to account for the appearance variation is to update the template every frame by taking the overlap between the CAD model and the current point cloud scan, shown in Fig. 4 (d). Nonetheless, this strategy still may not achieve ideal performance, as the template quickly becomes erroneous, as shown in Fig. 4 (e) and Fig.~\ref{fig:templatedrift}. This is because the point cloud noise causes the instability of the alignment and finally leads to a template mismatch.

\begin{figure}[h]
\begin{center}
\includegraphics[width=1.0\linewidth]{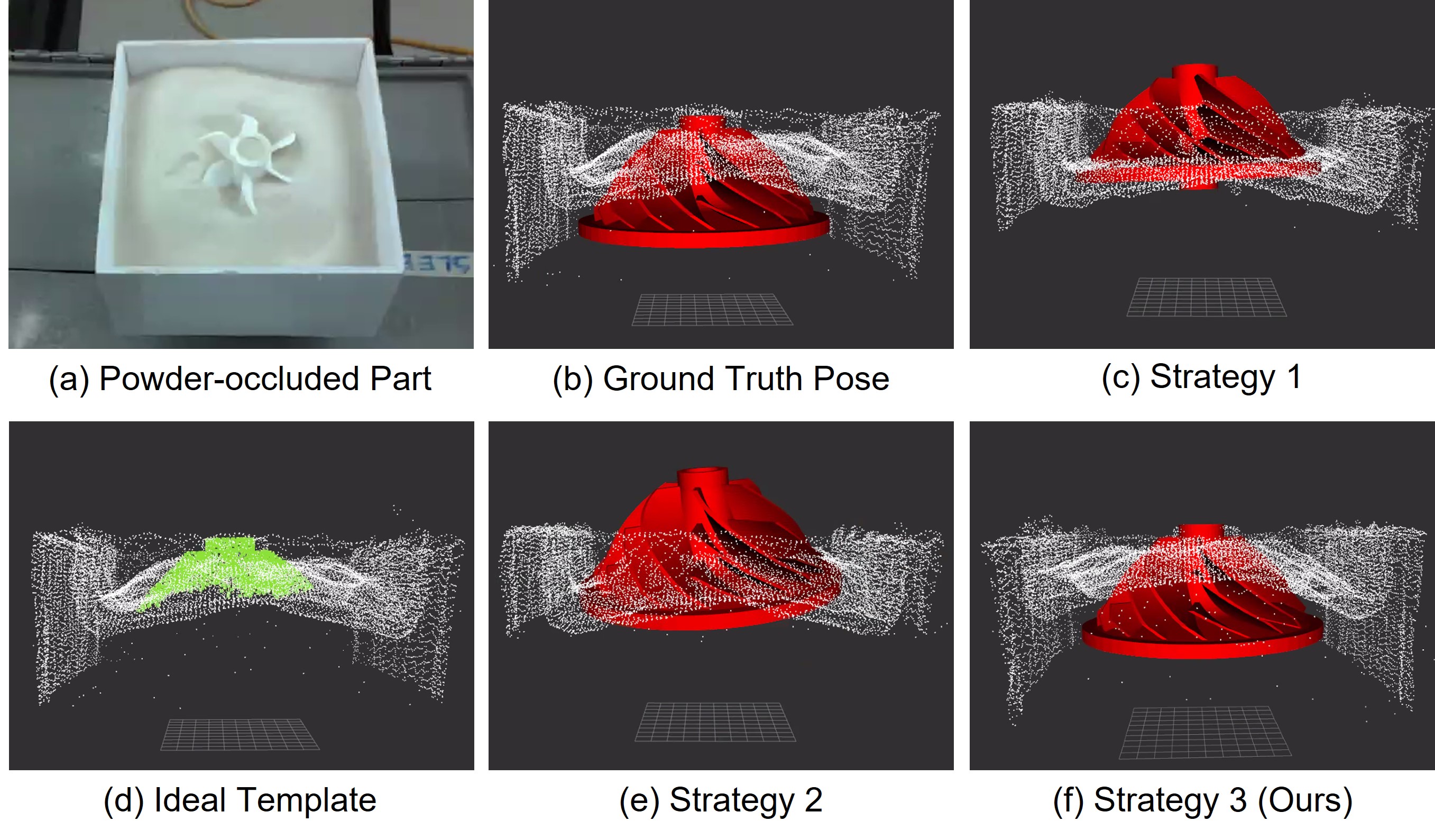}
\end{center}
\vspace{-15pt}
\caption{A comparison of different template update strategies. (a) A propeller partially covered by powder. (b) The CAD model rendered according to the ground truth pose. (c) Pose-tracking result with strategy 1. (d) Template taken from the overlapping area between the CAD model and the scan. (e),(f) Pose tracking results with Strategy 2 and our strategy. With Strategies 1 and 2, the bottom of the propeller is mistakenly aligned to the powder surface, resulting in incorrect estimated pose. Our strategy resolves this issue.}
\vspace{0pt}
\label{fig:comptemplate}
\end{figure}

\begin{figure}[h]
\begin{center}
\includegraphics[width=1.0\linewidth]{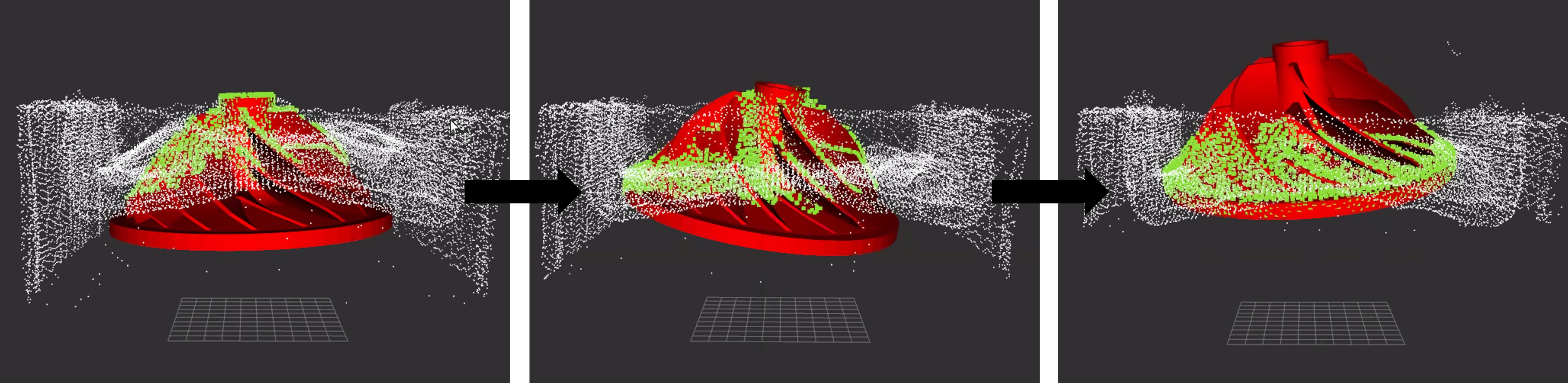}
\end{center}
\vspace{-10pt}
\caption{Visualization of template error accumulation with Strategy 2. The initial template aligns well with the point cloud (left). As tracking starts, although the target remains stationary, the sensor noise causes the oscillation of the estimated pose. The oscillation results in the template error accumulation, leading to an incorrect template and pose estimate.}
\vspace{-10pt}
\label{fig:templatedrift}
\end{figure}

\subsection{Conditional Update ICP}
\label{sec:CU-ICP}
We propose a conditional template update strategy for ICP to avoid accumulated template update errors. Specifically, the template is updated only when the part appearance has changed significantly. In depowdering, this happens either when (1) more part surface becomes visible as more powder is cleaned or (2) 3D-printed parts have moved either by the depowdering tool or due to the loss of balance. Either scenario causes the powder distribution to change.
\newline\indent\textbf{Strategy 3} The template is updated only when the orientation $R$, translation $t$, or depowdering progress $\eta$ change by a significant amount:
\begin{equation} \label{eq:rule}
\begin{aligned}
    &\mbox{If } |R_i - R_{last}| > \delta_1 \mbox{ or } |t_i - t_{last}| > \delta_2 \mbox{ or } |\eta_i - \eta_{last}| > \delta_3, \\
    &\mbox{then } P_{tmpl}^{i} = (P_{cad} \cdot T_i) \cap P_{i} \\
    &\mbox{else } P_{tmpl}^{i} = P_{tmpl}^{i-1},
\end{aligned}
\end{equation}
where $R_{last}$, $t_{last}$, and $\eta_{last}$ are the orientation, translation, and progress at the last time when the template was updated, $P_{tmpl}^{i}$ is the current template, and $P_{cad} \cdot T_i$ is the CAD model transformed to the current pose. $\delta_1$, $\delta_2$, and $\delta_3$ are user-defined thresholds that control the template update frequency. Since this rule avoids unnecessary template update, the shape of the template is preserved throughout depowdering. 

With this new template update rule, we develop our \textit{Conditional Update ICP} (CU-ICP) algorithm shown in Alg.~\ref{alg:ATICP}. The template is initialized (Lines 4-6) by transforming the CAD model to the initial part pose and running Alg.~\ref{alg:templateupdate}. Then, ICP constantly registers the template to the current scan, updates the current pose, and transforms the CAD model to the new pose (Lines 8-10). Based on the new pose, the depowdering progress is calculated, which will be discussed further in Sec.~\ref{sec:progress}. If the condition is satisfied, the template will be updated (Lines 12-16). Specifically, for each point $p_{cad}$ in the CAD model, Alg.~\ref{alg:templateupdate} finds its nearest neighbor $p_{nb}$ from the scan (Line 3). If the distance is within the template update distance threshold, $\xi$, then $p_{cad}$ is considered to be matched, and this point is added to the updated template (Lines 5-6).

\begin{algorithm}
\caption{Conditional Update ICP}
\begin{algorithmic}[1]
\While{$true$}
\State $P_{i} \gets$ {\bf GetCurrentScan};
\If{$first$ $iteration$}
\State $T_{i} \gets T_{init}$;
\State $P_{cad} \gets$ {\bf TransformPointCloud}$(P_{cad}, T_{init})$;
\State $P_{tmpl} \gets$ {\bf TemplateUpdate}$(P_{cad}, P_{i})$;
\EndIf
\State $T_{i,i-1} \gets$ {\bf ICP}$(P_{tmpl}, P_{i})$;
\State $T_{i} \gets T_{i,i-1} \times T_{i}$;
\State $P_{cad} \gets$ {\bf TransformPointCloud}$(P_{cad}, T_{i,i-1})$;
\State $\eta_i\gets$ {\bf ProgressEstimation}$(P_{cad}, P_{i})$;
\If{$R_i - R_{last} >\delta_1$ || $t_i - t_{last} >\delta_2$ || $\eta_i - \eta_{last} > \delta_3$}
\State $P_{tmpl} \gets$ {\bf TemplateUpdate}$(P_{cad}, P_{i})$;
\State $R_{last} \gets R_{i}$;
\State $t_{last} \gets t_{i}$;
\State $\eta_{last} \gets \eta_{i}$;
\EndIf
\EndWhile
\end{algorithmic}
\label{alg:ATICP}
\end{algorithm}

\begin{algorithm}
\caption{Template Update}
\begin{algorithmic}[1]
\State $P_{tmpl} \gets$ \O
\For{each $p_{cad}$ in $P_{cad}$}
\State $p_{nb} \gets$ {\bf FindNearestNeighbor}$(p_{cad}, P_{i})$;
\State $d \gets$ {\bf CalcEuclideanDist}$(p_{cad}, p_{nb})$;
\If{$d < \xi$}
\State $Add$ $p_{cad}$ to $P_{tmpl}$;
\EndIf
\EndFor
\State {\bf return} $P_{tmpl}$;
\end{algorithmic}
\label{alg:templateupdate}
\end{algorithm}

\subsection{Progress Estimation}\label{sec:progress}
Progress estimation monitors the percentage of completion for depowdering, with 0\% meaning parts are completely covered, and 100\% meaning depowdering is completed. It is also an important factor for conditional template update, as described in Sec.~\ref{sec:CU-ICP}. Given an entire CAD model and an updated template that represents the visible portion, an intuitive way is to calculate the ratio between their surface area or volume. However, these two metrics may not accurately reflect the depowdering progress due to the potential non-linear changing pattern for 3D-printed parts whose cross section area varies significantly in the vertical direction. In fact, the depowdering progress is directly related to the powder surface height, because the unfused powder is removed layer by layer with a constant thickness drop through vacuuming. Therefore, we select the height ratio between the template and the CAD model to reflect the depowdering progress.

\begin{figure}
\begin{center}
\includegraphics[width=1.0\linewidth]{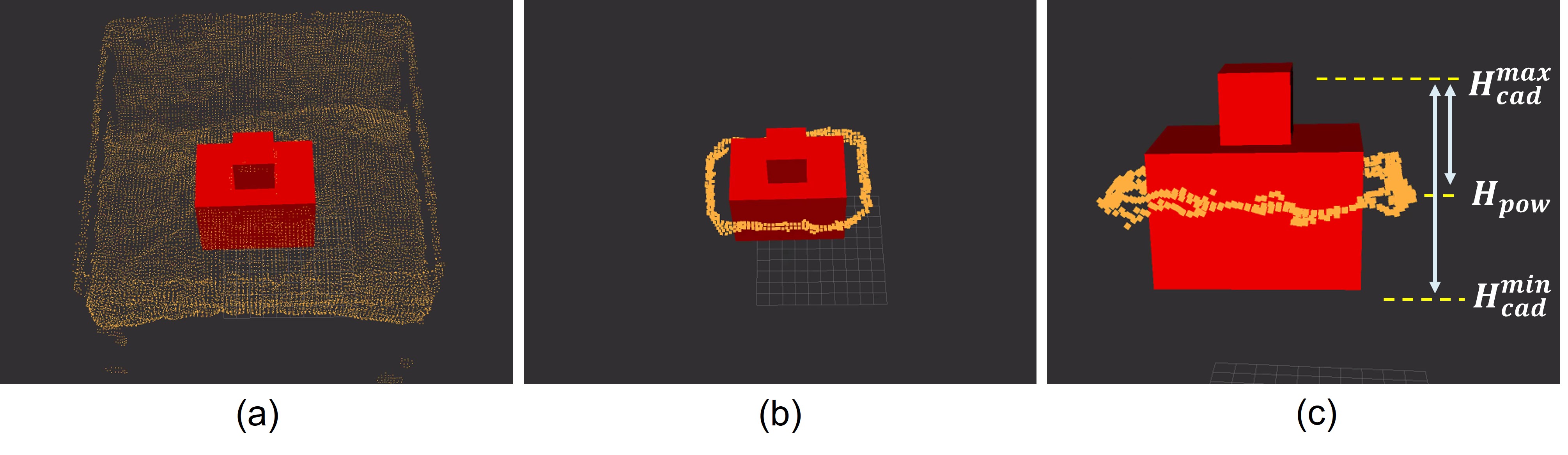}
\end{center}
\vspace{-15pt}
\caption{Powder contour extraction and progress estimation. (a) Powder (orange) segmented from the scan. (b) Powder contour generated by selecting points surrounding the part contour. (c) Progress estimation based on the height ratio of the visible surface and the whole part.}
\vspace{-10pt}
\label{fig:progress}
\end{figure}

To estimate the depowdering progress based on height ratio, the first step is to identify the powder contour around the visible part surface. As shown in Fig.~\ref{fig:progress} (a), based on the estimated pose, the point cloud scan can be segmented into part and powder. The powder contour is then extracted by selecting powder points that are certain distances away from the part segment\footnote{In scenarios where the 3D-printed part is fully covered, the segmentation can still be performed, as the points in the scan that are close to the top of the part will be grouped as the part segment. Therefore, the powder contour can still be generated based on the segmentation result at this stage.}, as shown in Fig.~\ref{fig:progress} (b). By calculating the average height of the powder contour $H_{pow}$, the depowdering progress\footnote{The depowdering progress $\eta$ is always estimated in real-time based on the current pose, even if the pose has changed in the middle of depowdering. The robot will keep removing powder until $\eta$ reaches the target number.} $\eta$ can be calculated by:
\begin{equation}\label{eq:eta}
   \eta = \frac{H_{cad}^{max} - H_{pow}}{H_{cad}^{max} - H_{cad}^{min}},
\end{equation}
where $H_{cad}^{max}$ is the height of the CAD model's top surface and $H_{cad}^{min}$ is the height of the CAD model's bottom surface. If $H_{pow}$ is higher than $H_{cad}^{max}$, $H_{pow}$ is truncated to $H_{cad}^{max}$.

\section{Experiments}
In this section, we first introduce the experimental setup. Then, we perform several experiments to validate the tracking performance of our CU-ICP. Finally, we present several complete demonstrations of the overall depowdering processes achieved by the robotic depowdering system. 

\subsection{Experimental Setup}
\begin{figure}
\begin{center}
\includegraphics[width=1.0\linewidth]{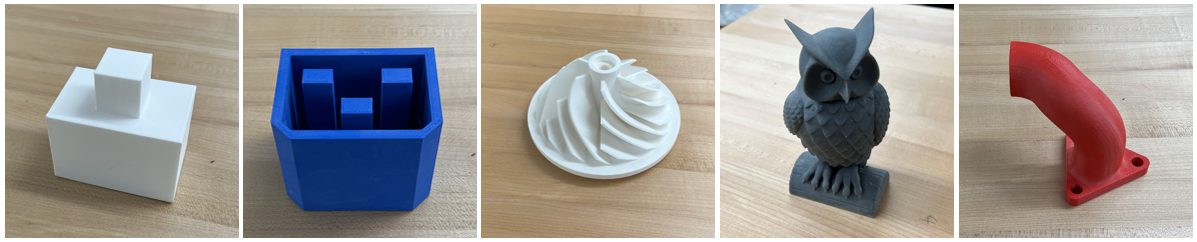}
\end{center}
\vspace{-10pt}
\caption{Parts selected for evaluation. From left to right are: cube, cup, propeller, owl, and pipe. Although these parts are printed in different colors, the color information is not used.}
\vspace{-10pt}
\label{fig:evalparts}
\end{figure}

\noindent\textbf{Evaluation Objects}. We evaluate our system on five 3D-printed parts with different shapes: cube, cup, propeller, owl, and pipe, shown in Fig.~\ref{fig:evalparts}. The sizes of these objects, defined as the diagonal length of the object's minimum 3D bounding box, vary from 8 centimeters to 15 centimeters. We cover parts with various amounts of powder to generate a range of surface visibility. For convenience, we define surface visibility the same as depowdering progress, see Eq.~\eqref{eq:eta}.

\noindent\textbf{Evaluation Metrics}. For stationary objects, we report the following metrics for pose tracking: 1) $\mathbf{R}_\mathrm{err}$: mean rotation error in degrees, and 2) $\mathbf{t}_\mathrm{err}$: mean translation error in centimeters. For moving objects, we employ the success rate and the maximum trackable speed. 

\noindent\textbf{Baseline Comparison}. We compare our algorithm with two baselines: (1) Vanilla ICP, where the entire CAD model is used as a template and no template update is performed; (2) Continuous ICP, where the template is updated in every frame based on the method specified by Alg.~\ref{alg:templateupdate}.

\noindent\textbf{Key Parameters}. For depowdering progress thresholds, we select $\eta_1 = 30\%$ and $\eta_2 = 85\%$, as mentioned in Sec.~\ref{sec:architecture}. For template update frequency thresholds, we empirically select $\delta_1 = 30\mathrm{deg}$, $\delta_2 = 5\mathrm{cm}$, and $\delta_3 = 15\%$, as we notice that an overly large threshold leads to an outdated template caused by the less frequent template update, while an overly small threshold ends with very frequent update that results in error accumulation as explained in Sec.~\ref{sec:mismatch}. For the template update distance threshold, we empirically select $\xi$ = 1cm for cube, cup, and owl. For propeller and pipe, since their shapes are more complicated, which causes a noisier scanned point
cloud, we conservatively select $\xi$ = 0.8cm to avoid more outliers being included into the template.

\subsection{Pose Tracking for Static Objects}
\begin{figure}[h]
\begin{center}
\includegraphics[width=1.0\linewidth]{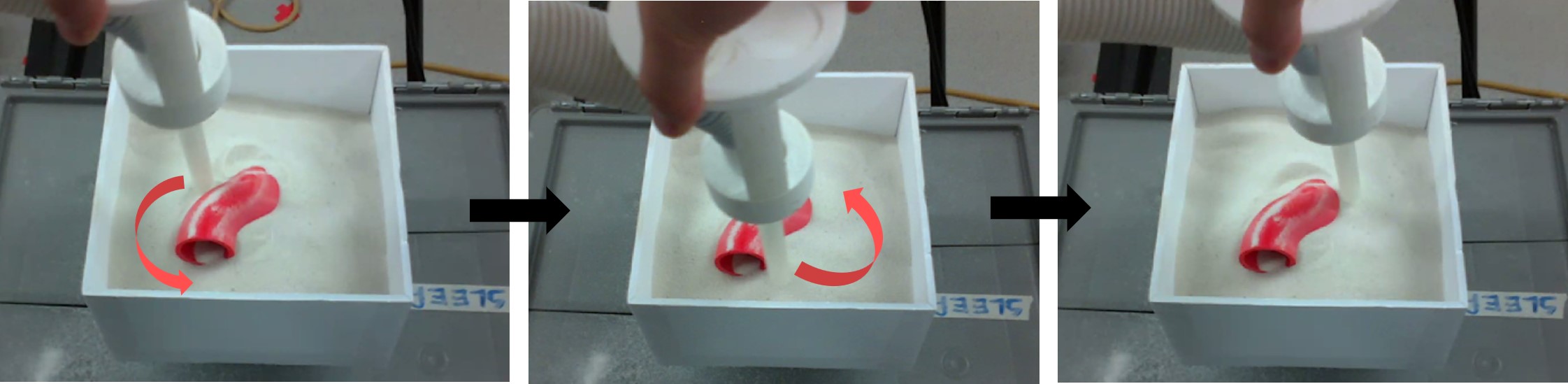}
\end{center}
\vspace{-10pt}
\caption{Experimental setup for tracking static objects at different visibility. We manually move the nozzle around parts to simulate the actual depowdering environment with point cloud occlusion and interference. The red arrow shows the nozzle moving direction.}
\label{fig:statictrack}
\end{figure}

In a real depowdering environment, 3D-printed parts are partially visible. The robot end-effector introduces extra interference and occlusion while moving around parts. To test the pose-tracking accuracy under this condition, we manually move a nozzle around stationary parts, with the visibility ranging from 20\% to 100\%, as shown in Fig.~\ref{fig:statictrack}. 

\begin{table*}
\begin{center}
\begin{minipage}[b]{.9\textwidth}
\caption{Pose tracking error with different levels of surface visibility}
\label{tab:trackerr} 
\resizebox{\textwidth}{!}{
\begin{threeparttable}
\begin{tabular}{c|c|ccccc|ccccc}
\toprule
\multirow{2}{*}{} & &\multicolumn{5}{c|}{$\bf R_{err}(degree)$} & \multicolumn{5}{c}{$\bf t_{err}(cm)$}\\
& Surface Visibility & 20\% & 40\% & 60\% & 80\% & 100\% & 20\% & 40\% & 60\% & 80\% & 100\%\\
\midrule
\multirow{6}{*}{CU-ICP (ours)} & Cube & 16.8 & $\bf 1.69$ & $\bf 0.73$ & $\bf 2.87$ & $\bf 4.29$ & $\bf 1.05$ & $\bf 0.15$ & $\bf 0.19$ & 0.28 & 0.35\\[2pt]
& Cup & $\bf 3.85$ &$\bf 4.05$ & $\bf 3.76$ & $\bf 3.06$ & $\bf 5.03$ & $\bf 0.94$ & $\bf 1.07$ & $\bf 0.68$ & $\bf 0.67$ & $\bf 0.60$ \\[2pt]
& Propeller & $\bf 6.60$ & $\bf 2.62$ & $\bf 2.12$ &$\bf 2.26$ & $\bf 2.45$ & $\bf 0.83$ & $\bf 0.39$ &$\bf 0.36$ & $\bf 0.15$ & $\bf 0.19$\\[2pt]
& Owl &$\bf 47.9$ & $\bf 1.55$ & 12.4 & $\bf 16.4$ & $\bf 5.85$ & $\bf 2.20$ & $\bf 1.55$ & $\bf 0.66$ & $\bf 0.80$ & $\bf 0.40$ \\[2pt]
& Pipe & $\bf 19.6$ & $\bf 8.53$ & $\bf 6.74$ & $\bf 5.35$ & $\bf 3.93$ & $\bf 1.17$ & $\bf 0.30$ & $\bf 0.38$ & $\bf 0.32$ & $\bf 0.34$\\[2pt]
& Overall & $\bf 19.0$ & $\bf 3.69$ & $\bf 5.15$ & $\bf 5.99$ & $\bf 4.31$ & $\bf 1.24$ & $\bf 0.69$ & $\bf 0.45$ & $\bf 0.44$ & $\bf 0.38$\\
\midrule
\multirow{6}{*}{Continuous ICP}& Cube & 16.8 & 3.76 & 1.80 & 3.77 & 4.46 & 1.38 & 0.22 & 0.24 & $\bf 0.27$ & 0.35\\[2pt]
& Cup & 50.3 & 36.1 & 34.3 & 13.4 & 5.66 & 6.54 & 2.08 & 1.13 & 0.85 & 0.79 \\[2pt]
& Propeller & 11.1 & 12.7 & 4.04 & 12.7 & 10.8 & 4.14 & 2.88 & 1.70 & 0.70 & 0.49 \\[2pt]
& Owl & 74.0 & 39.1 & $\bf 12.3$ & 26.7 & 7.01 & 4.79 & 2.66 & 0.79 & 1.31 & 0.42\\[2pt]
& Pipe & 55.5 & 29.7 & 18.5 & 6.79 & 4.16 & 2.32 & 1.25 & 1.38 & 0.60 & 0.38 \\[2pt]
& Overall & 41.5 & 24.3 & 14.2 & 12.7 & 6.42 & 3.83 & 1.82 & 1.05 & 0.75 & 0.49\\[2pt]
\midrule
\multirow{6}{*}{Vanilla ICP}& Cube & 18.1 & 3.52 & 2.43 & 3.49 &  4.49 & 1.41 & 0.23 & 0.27 & 0.30 & 0.35\\[2pt]
& Cup & 42.0 & 36.3 & 36.0 & 13.3 & 5.80 & 2.94 & 2.04 & 1.15 & 0.91 & 0.80 \\[2pt]
& Propeller & 10.4 & 7.61 & 4.47 & 16.9 & 16.2 & 4.02 & 2.85 & 1.78 & 0.62 & 0.44\\[2pt]
& Owl & 75.1 & 69.9 & 27.0 & 28.9 & 6.75 & 4.91 & 4.64 & 1.69 & 1.40 & 0.41\\[2pt]
& Pipe & 57.2 & 30.0 & 18.9 & 6.86 & 4.25 & 2.38 & 1.30 & 1.14 & 0.61 & 0.38\\[2pt]
& Overall & 40.6 & 29.5 & 17.8 & 13.9 & 7.50 & 3.13 & 2.21 & 1.21 & 0.77 & 0.48\\[2pt]
\bottomrule
\end{tabular}
\begin{tablenotes}
\footnotesize
\item[1] Surface visibility here is defined the same as depowdering progress.
\item[2] $R_{err}$ and $t_{err}$ represent rotation error and translation error, respectively. The lower the better.
\end{tablenotes}
\vspace{-10pt}
\end{threeparttable}}
\end{minipage}
\end{center}
\end{table*}

As shown in Table~\ref{tab:trackerr}, our algorithm achieves significant better performance compared to Continuous ICP and vanilla ICP, especially when the surface visibility is less than or equal to 60\%. The better performance with additional point cloud occlusion and interference shows the effectiveness of the template update criteria.
In fact, in this static object scenario where no significant changes occur on either $\eta$ or $T_i$, CU-ICP does not perform unnecessary template update and therefore is able to maintain a stable template.

\begin{figure}
\begin{center}
\includegraphics[width=1.0\linewidth]{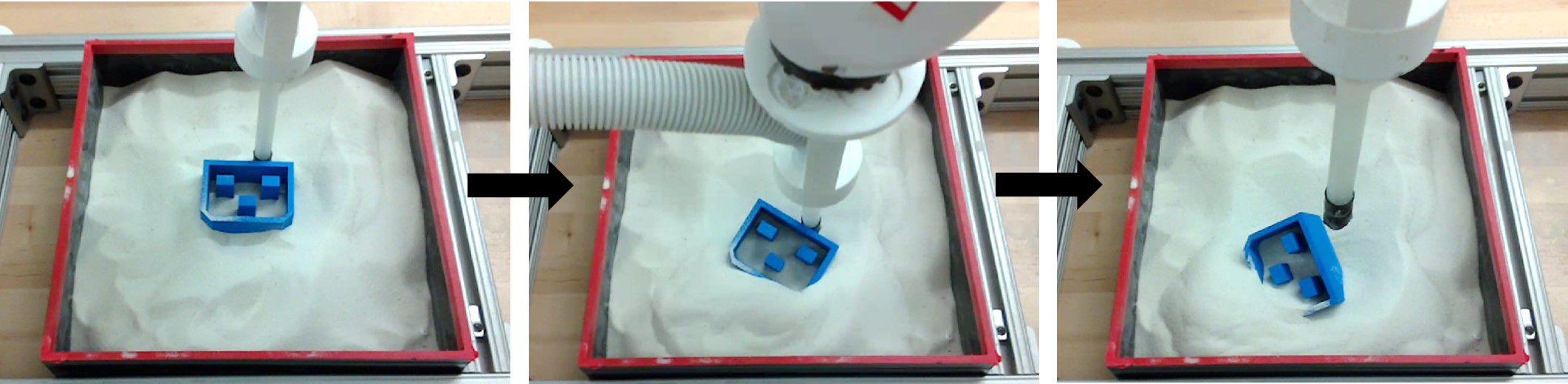}
\end{center}
\vspace{-10pt}
\caption{Experimental setup for tracking moving objects, with the initial surface visibility ranging from around 40\% to 60\%. We use the robotic arm to push the part until it reaches the box edge, and test if the part pose can be tracked during the motion.}
\label{fig:movingvis}
\end{figure}

\begin{table}[t]
\centering
\begin{minipage}[b]{.49\textwidth}
\centering
\caption{Success rate (\%) of tracking moving objects in powder \label{lab}}
\label{tab:successrate} 
\resizebox{\textwidth}{!}{
\begin{tabular}{|c|c|c|c|c|c|c|}
\hline
Method & Cube & Cup & Propeller & Owl & Pipe & Overall \\
\hline
\bf CU-ICP (ours) & 100 & \bf{80} & \bf{100} & 100 & \bf{80} & \bf{92} \\
\bf Continuous ICP & 100 & 0 & 40 & 100 & 20 & 52\\
\bf Vanilla ICP & 60 & 0 & 0 & 100 & 0 & 32\\
\hline
\end{tabular}
}
\vspace{-10pt}
\end{minipage}
\end{table}

\begin{figure*}[h]
\begin{center}
\includegraphics[width=1.0\linewidth]{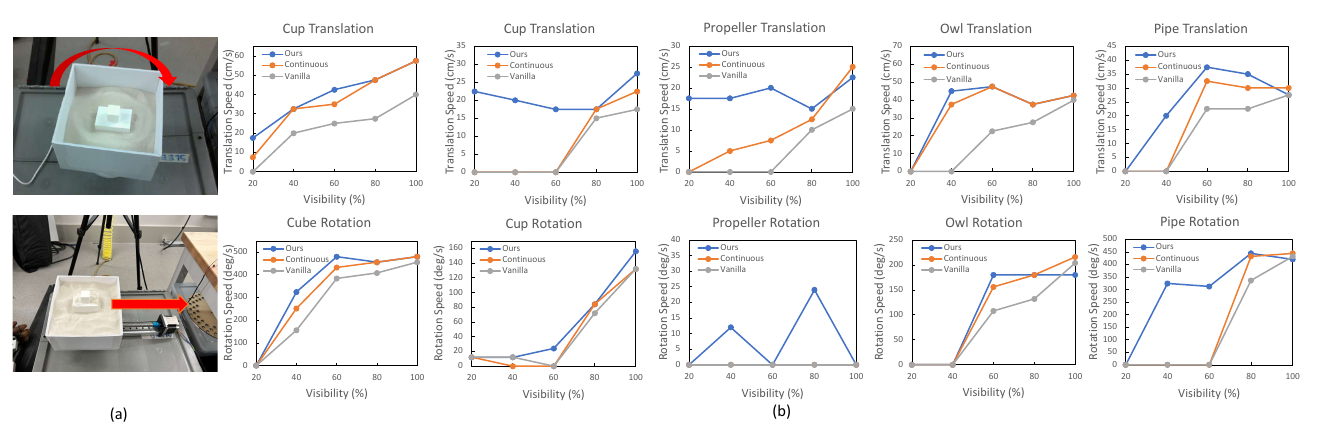}
\end{center}
\vspace{-20pt}
\caption{(a) Experimental setup for maximum trackable moving speed test. We use a linear actuator and a turntable to generate translation and rotation for 3D-printed parts. (b) Maximum trackable moving speed in translation and rotation for different 3D-printed parts with various surface visibility.}
\vspace{0pt}
\label{fig:speedvis}
\end{figure*}

\begin{figure*}
\begin{center}
\includegraphics[width=1.0\linewidth]{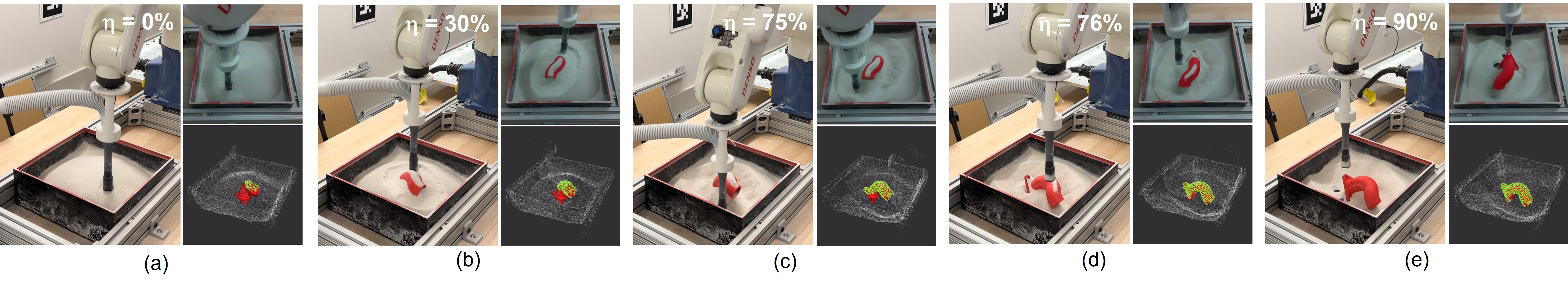}
\end{center}
\vspace{-15pt}
\caption{A complete depowdering process for the pipe. The lower right picture of each snapshot shows the estimated pose in red and the corresponding template in green. (a) Progress is 0\%. The pipe is entirely covered by powder. (b) Progress reaches 30\%. CU-ICP starts tracking the part pose. (c) The part becomes more visible, and the robot continues to remove powder along the outer contour of the pipe. (d) The pipe loses balance due to the loss of powder support. The robot adjusts to the new pose and avoids collision. (e) The robot removes the residual powder through air blasting.}
\vspace{-5pt}
\label{fig:demopipe}
\end{figure*}

\begin{figure*}
\begin{center}
\includegraphics[width=1.0\linewidth]{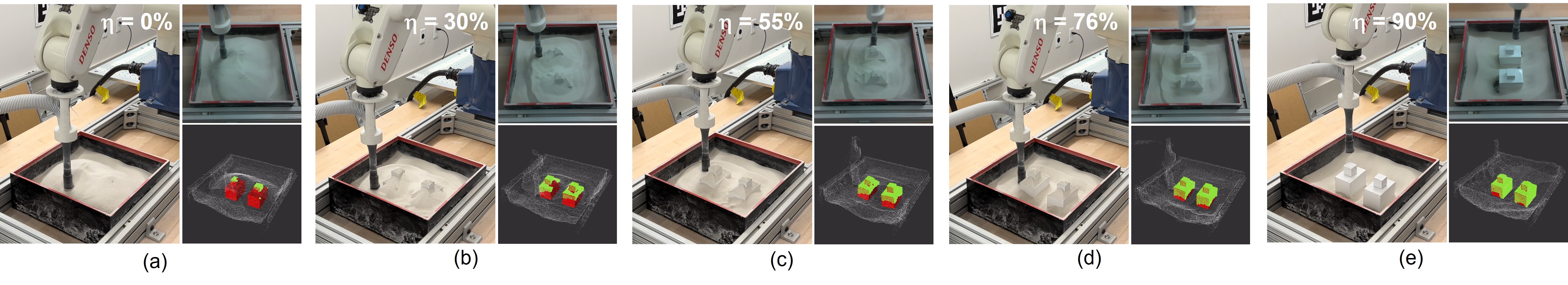}
\end{center}
\vspace{-15pt}
\caption{A complete depowdering process for two cubes. We run two tracking processes in parallel to improve tracking efficiency. The robot first performs depowdering on the left cube for one cycle, and then moves on to the right cube for another cycle. (a) Progress is 0\%. The cubes are entirely covered by powder. (b) Progress reaches 30\% and two cubes starts being tracked in parallel. (c)-(d) CU-ICP keeps tracking poses and two cubes gradually become more visible. (e) The residual powder is removed through air blasting.}
\vspace{-15pt}
\label{fig:democube}
\end{figure*}

\subsection{Pose Tracking for Moving Objects}
When a 3D-printed part moves through powder, the powder distribution changes as the powder piles up on one side and spreads out on the other side. To examine the tracking performance for moving objects, we use the robotic arm to push parts at different initial contact points, till they reach the edge of the build box. The resulting motion is a combination of translation and rotation. With the initial pose given, we run pose tracking while parts are moving. For each selected part in Fig.~\ref{fig:evalparts}, we randomly sample five initial contact points. Hence, 25 trials are performed in total. A trial is defined as successful if the final orientation error is within 15deg and the final translation error is within 15mm.

As shown in Table~\ref{tab:successrate}, the overall success rate of CU-ICP is significantly higher than those of other baselines. Especially, both Continuous ICP and Vanilla ICP produce low success rate on propeller, cup, and pipe, while CU-ICP shows a better performance among all different parts. The reason is that our conditional template update is robust to the inconsistency between the point cloud scan and the template, which is caused by either invisible inner structures of 3D-printed parts or the limited camera capability. For example, the cup and propeller contain detailed inner structures that are difficult to be captured by the 3D cameras; the pipe contains surface curvatures that cannot be accurately reconstructed due to the limitation of cameras. Without the conditional template update for maintaining accurate shapes of the templates, the inconsistency easily leads to pose-estimation errors and template mismatch. The fact that our CU-ICP is less affected by these drawbacks demonstrates its robustness to point cloud occlusion as well as point cloud distortion.

\subsection{Maximum Trackable Speed}
We test the maximum part moving speed at which the tracking algorithms lose track of the selected parts. We use a linear actuator and a turntable to generate adjustable translation and rotation\footnote{Due to the hardware limit, the maximum moving speed that the linear actuator and turntable can generate is limited. Therefore, in order to achieve arbitrary moving speed, we playback the recorded point cloud sequence at faster frame rates instead of physically moving parts. Then we uniformly downsample the sequence to the normal frame rate.}, as shown in \fref{fig:speedvis} (a). 

As shown in \fref{fig:speedvis} (b), it is obvious that CU-ICP results in higher maximum trackable speed compared with Continuous ICP and Vanilla ICP with visibility of less than 80\%. As the visibility increases, the maximum trackable speeds for the three algorithms gradually become close to each other. This is because the updated template is more similar to the original CAD model. Also, note that the maximum trackable speed for CU-ICP and Continuous ICP may not strictly increase along with the visibility, such as the cup translation with visibility from 20\% to 80\% and the pipe translation with visibility from 40\% to 100\%. This indicates that increasing target visibility does not necessarily result in a higher trackable speed. In fact, as the target visibility increases, the number of points in the template to be processed also increases, which slows down the computation and reduces the tracking capability of the algorithms. Overall, CU-ICP can be run on a laptop CPU with the maximum processing rate of around 60 FPS. 
\vspace{-5pt}

\subsection{Depowdering with the Robotic System}
Finally, we demonstrate several complete depowdering processes achieved by our robotic system. Two depowdering scenarios are considered here: (1) single-part depowdering, and (2) multiple-part depowdering, as shown in Fig.~\ref{fig:demopipe} and Fig.~\ref{fig:democube}. 
In the former scenario, the 3D-printed part loses balance during the process and falls to one side. 
The robot is able to adjust to the new pose and successfully avoids collision. 
In the latter scenario, the pose tracking for different 3D-printed parts is running in parallel to improve the computation efficiency. The robot successfully removes powder for all parts. The depowdering time depends on the thickness of the powder, the robot moving speed, as well as the vacuuming power. With the current experimental setup, the first process takes 3min 50sec, and the second process takes 3min 47sec. The two demonstrations prove that our system is able to automate depowdering for parts with various shapes without the concern of damaging fragile parts. 

\vspace{-0pt}
\section{Conclusions And Future Directions}
This paper proposes a robotic system that automatically removes unfused powder from the surface of 3D-printed parts using visual feedback.
The main component is a visual perception system that takes in the scanned point cloud from the depth camera, tracks the 6D pose of powder occluded parts, and identifies the powder contour for progress estimation and path generation. Experiments show that the visual perception system is robust to occlusion and sensor noise. In particular, our tracking algorithm, named Conditional Update ICP (CU-ICP), achieves higher tracking performance compared to other baselines. The robotic system is able to automate depowdering for parts with various shapes without the the need for pre-depowdering and the concern of damaging fragile parts. 
\newline\indent In future work, we aim to further remove the residual powder remaining inside 3D-printed parts by investigating more advanced path planning strategies and combining various cleaning approaches, in order to achieve a higher level of depowdering automation.

\vspace{-5pt}
\section*{ACKNOWLEDGMENT}
We are grateful to Dr. Chen Wang at the Robotics Institute, Carnegie Mellon Univeristy, for his advice and support to the research.
This work was funded by Carnegie Mellon University Manufacturing Futures Initiative and NASA University Leadership Initiative.
\vspace{-5pt}
\bibliographystyle{IEEEtran}
\bibliography{references}

\end{document}